# Statistical Sign Language Machine Translation: from English written text to American Sign Language Gloss

Achraf Othman[1] and Mohamed Jemni[2]

Research Lab. UTIC, University of Tunis
5, Avenue Taha Hussein, B. P. : 56, Bab Menara, 1008 Tunis, Tunisia

**Abstract**
This works aims to design a statistical machine translation from English text to American Sign Language (ASL). The system is based on Moses tool with some modifications and the results are synthesized through a 3D avatar for interpretation. First, we translate the input text to gloss, a written form of ASL. Second, we pass the output to the WebSign Plug-in to play the sign. Contributions of this work are the use of a new couple of language English/ASL and an improvement of statistical machine translation based on string matching thanks to Jaro-distance.

## 1. Introduction

For many centuries, Deaf have been ignored, considered mentally ill. And there wasn't effort to try to contact them. Only close deaf communicated with each other. In the 18th century, deaf people are beginning the use of a Sign Language (SL) based on gestural actions. Gestures that can express human thought as much as a spoken language. This gestural language was not a real methodical language what follows an anarchic development of sign language for a long time. Within seventies that hearing persons wishing to learn the language of the deaf and the deaf willing to teach find themselves in the school to learn. It is therefore necessary to develop teaching materials and accessible educational tools. It is very unfortunate but there is no universal sign language, each country has its own sign language. Communication between the deaf and hard of hearing from different countries or community is a problem, knowing that most deaf people do not know how to read or write. From the Eighties, researchers begin to analyze and process sign language. Next, they design and develop routines for communication intra-deaf and between hearing and deaf people. Starting from the design of automatic annotation system of the various components of sign language and coming to the 3D synthesis of signs through virtual avatars. In recent years, there was the appearance of a new line of research said automatic Sign Language Processing noted SLP. SLP is how to design, represent and process sign language incompletely described [1]. After that, there was the appearance of some works towards translate automatically written text to sign language. There are two types of machine translation. First, those which generate a 3D synthesis through a virtual character who plays the role of an interpreter in sign language. Second, those generate glosses from written text. Usually, any automatic processing of language (natural or signed) requires corpus to improve treatment outcomes. Note that sign languages are made up of manuals components and non-manual components such as gaze, facial expressions and emotions. The purpose of this paper is to focus on how to use statistics to implement a machine translation for sign language. This paper begins with an overview of various kind of machine translation for sign language. Next, an overview of our contribution and structure is introduced. Section 4 is a short description of the parallel data English/ASL. Alignment and training steps are shown in section 5. Section 6 describes our contribution in statistical machine translation. Phrase-base model and decoding are explained in section 7. Results and word alignment matrix are illustrated in section 8. Conclusions are described in section 9.

## 2. Machine Translators to Sign Language

Machine translators have become more reliable and effective through the development of methodology for calculating and computing power. The first translators appeared in the sixties to translate Russian into English. The first machine translator considered this task as a phase of encryption and decryption. Today and following technological developments, there was appearance of new systems. Some based on grammatical rules and other based on statistics. Not forgetting that there are translators which are based on examples. The translation stage requires preprocessing of the source language as sentence boundary detection, word tokenization, chunking… And, these treatments requires corpus. After the evolution of corpora size of and diversity for written language, there were multitudes of machine translators for the majority of languages in the world. But for sign language, we found only a few projects that translate a textual language to sign language or sign language to written text. In what follows we present various existing projects for sign language.

2.1 TEAM Project

TEAM [2] was an English-ASL translation system. It built at the University of Pennsylvania that employed synchronous tree adjoining grammar rules to construct an ASL syntactic structure. The output of the linguistic portion of the system was a written ASL gloss notation system [3] with embedded parameters. This notation system encoded limited information about morphological variations, facial expressions, and sentence mood. For synthesis, the authors took advantage of the virtual human modeling research by using an animated virtual character as signing avatar. The project had particular success at generating aspectual and adverbial information in ASL using emotive capabilities of the animated character.

2.2 English to American Sign Language: Machine Translation of Weather reports

As is common with machine translation systems, the application [4] consists of four components: a lexical analyzer, a parser, a transfer module and a generation module. In addition, there is an initial module that obtains the weather reports from the World Wide Web. Several of the components use freely available Perl modules, packages designed to assist in those particular tasks for spoken or computer languages. The ASL generation module uses the notion of "sentence stems" to generate fluent ASL. The Perl script first takes an inventory of the kinds of information present in the semantic representation, and generates a formulaic phrase for each one. These formulas all use ASL grammar, including topic-comment structure and non-manual grammatical morphemes. The content that is output by the transfer module is then plugged in to the formulas, producing fluent ASL.

2.3 The South African Sign Language Machine Translation

The aim of the South African Sign Language Machine Translation (SASL-MT) project [5] is to increase the access of the Deaf community to information by developing a machine translation system from English text to SASL for specific domains where the need is greatest, such as clinics, hospitals and police stations, providing free access to SASL linguistic data and developing tools to assist hearing students to acquire SASL. The system reuses the same concept of TEAM Project [2]. So, authors constructed SASL grammar rules, and rule-based transfer rules from the English trees to SASL trees. These trees were built manually from a set of sentences. The system transferred all pronouns detected in the sentence to objects. Then, it placed them into signing space.

This project is still under development. The authors have completed the tag parser for the English, the metadata generator for pronoun resolution and generation of emotional, stress and accent flags, and the signing avatar. Also, there aren't experimental results.

2.4 Multipath-architecture for SL MT

Huenfaurth [6] described a new semantic representation that uses virtual reality scene. The aim of his work was to produce spatially complex American Sign Language (ASL) phenomena called "Classifier Predicates" [7]. The model acted as an Interlingua within new multi-pathway machine translation architecture. As opposed to spoken and written languages, American Sign Language relied on the multiple simultaneous channels of hand shape, hand location, hand/arm movement, facial expression and other non-manual gestures to convey the meaning. For this reason, the author used a multi-channel architecture to express additional meaning of ASL.

2.5 Czech Sign Language Machine Translation

The goal of this project was to translate spoken Czech to Signed Czech [8]. The system included a synthesis of sign by the computer animation. The synthesis employed a symbolic notation HamNoSys [9]. An automatic process of synthesis generated the articulation of hands from the notation. The translation system has built in the statistical ground. The inventory of Czech Sign Language used for the creating of complete vocabulary of signs. This dictionary had more than 3000 simple or linked signs and covers the fundamental vocabulary of Czech Deaf community.

2.6 ViSiCAST Translator

Marshall et al. at the University of East Anglia implemented a system for translating from English text into British Sign Language (BSL) [10]. Their approach used the CMU Link Parser to analyze an input English text. And they used Prolog declarative clause grammar rules to convert this linkage output into a Discourse Representation Structure. During the generation half of the translation process, Head Driven Phrase Structure rules are used to produce a symbolic SL representation script. This script is in the system's proprietary 'Signing Gesture Markup Language (SiGML)', a symbolic coding scheme for the movements required to perform a natural Sign Language.

2.7 ZARDOZ System

The ZARDOZ system [11] was a proposed English-to-Sign-Languages translation system using a set of hand-coded schemata as an Interlingua for a translation component. Some of the researches focused of this system were the use of artificial intelligence knowledge representation, metaphorical reasoning, and blackboard system architecture; so, the translation design is very knowledge and reasoning heavy. During the analysis stage, English text would undergo sophisticated idiomatic concept decomposition before syntactic parsing in order to fill slots of particular concept/event/situation schemata. The advantage of the logical propositions and labeled slots provided by a schemata-architecture was that commonsense and other reasoning components in the system could later easily operate on the semantic information.

## 2.8 Environment for Greek Sign Language Synthesis

The authors [12] present a system that performs SL synthesis in the framework of an educational platform for young deaf children. The proposed architecture is based on standardized virtual character animation concepts for the synthesis of sign sequences and lexicon-grammatical processing of Greek sign language (GSL) sequences. A major advantage of the proposed architecture is that it goes beyond the usual single-word approach which is linguistically incorrect, to provide tools to dynamically construct new sign representations from similar ones. Words and phrases are being processed and the resulting notation subset of a lexical database, HamNoSys [9] eventually transformed into GSL and animated on the clients' side via an H|Anim compliant avatar.

## 2.9   Thai - Thai Sign Machine Translation

The authors [13] propose a multi-phase approach, Thai-Thai Sign Machine Translation (TTSMT), to translate Thai text into Thai Sign language. TTSMT begins the translation process by segmenting the input sentence since Thai is a non-word boundary language, converting the segmented sentence into simple sentence forms since most Thai Sign are expressed in a sequence of such form, and then generating the intermediate sign codes which link a Thai word to its corresponding Thai Sign. The most appro-priate sign codes will be selected and rearranged in the spatial grammatical order for generating the Sign language with pictures. The distinction between the Thai text and Thai Sign Language in both grammar and vocabulary are concerned in each processing step to ensure the accuracy of translation. The developed system was implemented and tested to translate Thai sentences used in everyday life.

In this section, we talked about several projects aiming to translate written text to sign language. In what follows, we introduce our contribution.

## 3. Contribution and structure

3.1 Problematic

American Sign Language has emerged as the most structured Sign Language in the World. More than 20 countries, that their deaf communities sign ASL. In USA, the community of Deaf counts between one and two millions that uses ASL for communication. Deaf people can't read or write English. This is the main problem in their life. Nowadays, Internet and any tool for communication are very important in our life. So, they are not accessible for Deaf. This work aims to design a machine translation for the pair English/ASL toward helping Deaf people. It will be very helpful for interpreters, hearing people and Deaf education.

3.2 Approach

Figure 1 describes the full process of our statistical machine translation for sign language between English /ASL. In the beginning we prepare our parallel data. In fact, this data is a simple file that contains a pair of sentences, one in English and the second one in ASL that is described in the next section. This pipeline is inspired from the work of Koehn, Och and Marcu [14]. For word alignment, we used the GIZA++ statistical word alignment toolkit. This tool extracted a set of high-quality word alignment from the original unidirectional alignments sets. We include in this step a string matching algorithm. For Statistical Machine Translation (SMT) Decoder, we use MOSES [15].

## 4. Parallel Data: Sign Language Corpus

A corpus is a scientifically prepared collection of examples of how a language is commonly used. A corpus can contain a large number of written texts, or recorded or filmed conversations. Such data collections are used to explore the usage of a language or to find out about the vocabulary and grammar of this language. Sign Language is characterized by its interactivity and multimodality, which cause difficulties in data collection and annotation. Our corpus is composed by a pair of sentences English vs. American Sign Language (ASL). They are stored in a text file. ASL is annotated by gloss. Glosses [3] are written words, where one gloss represents one sign. Additional markings provide further information, e.g. non-manual signs. Unfortunately no gloss standard exists, which results in inconsistent annotated corpora. Figure 2 is a short dialogue between two deaf peoples. The conversation is stored into a file and ready for training.

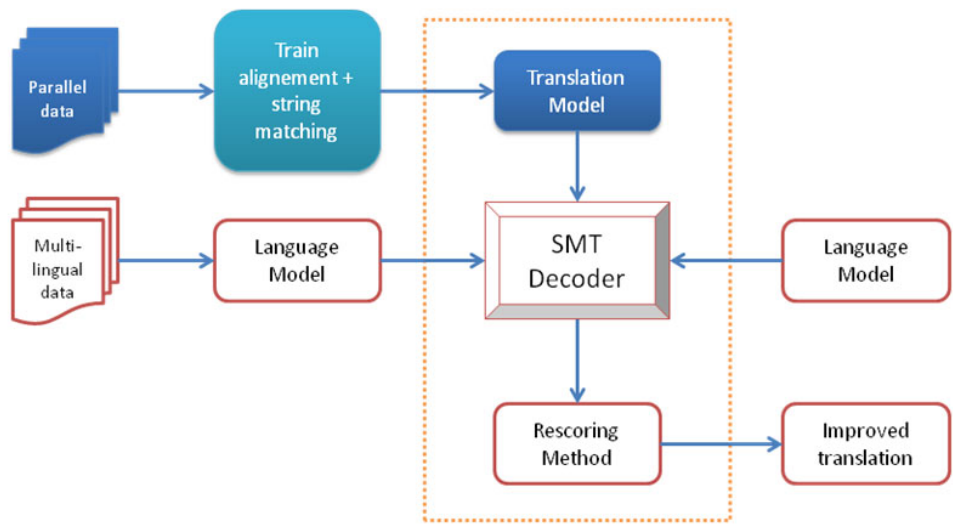

"Figure 1. Statistical Sign Language Machine Translation Pipeline"

**A: (get-attention) TOMORROW I GO PICK-up BOOK NEW I BUY YOU DON'T-MIND I BORROW YOUR TRUCK?**
*"Tomorrow I'm going to pick up some new books I just bought. Do you mind if I borrow your truck?"*
**B: TOMORROW TIME WHAT?**
*"What time tomorrow?"*
**A: AROUND 10 "give-or-take"**
*"Maybe around 10."*
**B: NO NOT WORK MY TRUCK me-BRING MECHANIC FIX TOMORROW MORNING. TOMORROW AFTERNOON BETTER.**
*"No, that won't work; I need to take the truck to get serviced tomorrow morning. The afternoon would work better."*
**A: FINE. YOU 2 TOMORROW FINE ?**
*"That's fine. Would 2 work for you?"*
**B: SURE-SURE FINE.**
*"Yes, that works fine."*

Figure 2. Example of conversation between 2 deaf peoples. Bold text is the ASL gloss and italic text is the English written version

## 5. Alignment and training

5.1. Word-based models

We present in this section a simple model for sign language machine translation that is based on lexical translation, the translation of words [15]. This method requires a dictionary that maps words from source language to target language, for example, from English to American Sign Language (ASL). If we take the word '*your*', we may find multiple translations to ASL like 'YOUR' or 'YOU'. Most words have multiple translations and some are more likely than others. For this reason, in some case, we cannot find the best translation if we use a dictionary to translate a sentence or a text. We refer to the use of statistics based on the count of words in a corpus or bilingual corpus. Table 1 displays the possible outcome of the word 'your'. This word occurs 148 times in our hypothetical text collection. It is translated 119 times into 'YOU' and 29 times in 'YOUR', and so on if there are other possible translations.

According to Koehn [15], we put formally the estimation of the lexical translation probability distribution from these counts. This function will returns a probability, for each choice of ASL translation e, that indicates how likely that translation is.

$$P_f = e \rightarrow P_f(e)$$

Table 1. Hypothetical counts for different translations of the English word 'your'

| Translation of 'your' | Count |
|---|---|
| YOU | 119 |
| YOUR | 29 |

Thanks to probability distribution for lexical translation, we can make a leap to our first model of statistical sign language machine translation, which use only lexical translation probabilities. We denote the probability of translating an English word $f$ into an ASL word $e$ with the conditional probability function $t(e|f)$. The alignment between input words and output words can be illustrated by a diagram:

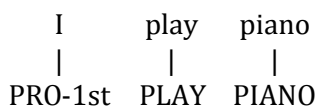

An alignment can be formalized with an alignment function a. This function maps, in our example, each ASL output word at position i to an English input word at position j: $a: j \rightarrow i$

This is a very simple alignment, since the English words and their ASL counterparts are in exactly the same order. While many languages do indeed have similar word order, a foreign language may have sentences in a different word order than is possible in ASL. This means that words have to be reordered during translation, as the following example illustrates:

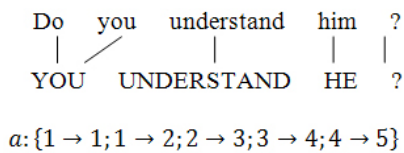

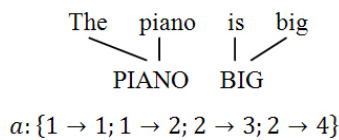

We have just laid some examples for alignment model based on words. Note that, in our alignment model, each output can be linked with one or more than input words, as defined by the alignment function. Several works implement this model, for example, the IBM Model for word alignment that is based on lexical translation probabilities [16]. There is 5 IBM Model for mapping words from a source language and a target language. For sign language machine translation and through the experimental results, we will implement only the three first models with an improvement algorithm based on string matching. In what follows, an implementation of three IBM Model is described.

### 5.2 IBM Model 1, 2 and 3

IBM Model 1 defines the translation probability for an English sentence $f = (f_1, \ldots, f_{l_f})$ of length $l_f$ to an ASL sentence $e = (e_1, \ldots, e_{l_e})$ of length $l_e$ with an alignment of each ASL word $e_j$ to an English word $f_i$ according to the alignment function $a: j \rightarrow i$ as follows:

$$p(e, a|f) = \frac{\epsilon}{(l_f + 1)^{l_e}} \prod_{j=1}^{l_e} t(e_j | f_{a(j)})$$

Let us take a look at how the algorithm words on a simple example. Table 2 presents a few iterations on a tiny three-sentence corpus with four input words (i, understand, play, piano) and four output words (I, UNDERSTAND, PLAY, PIANO). Initially, the translation probability distributions from the English words to the ASL words are ¼=0.25. Given this initial model, we collect counts in the first iteration of the EM algorithm. All alignments are equally likely.

Table 2. Application of IBM Model 1 EM Training: Given the three sentence pairs, the algorithm converges to values for t(e|f)

| e | f | initial | 1st it. | 2nd it. | 3rd it. | ... | Final |
|---|---|---|---|---|---|---|---|
| I | i | 0.25 | 0.41 | 0.53 | 0.64 | .. | 1.0 |
| I | understand | 0.25 | 0.50 | 0.45 | 0.38 | .. | 0.0 |
| I | play | 0.25 | 0.33 | 0.27 | 0.23 | .. | 0.0 |
| I | piano | 0.25 | 0.33 | 0.27 | 0.23 | .. | 0.0 |
| PIANO | i | 0.25 | 0.16 | 0.12 | 0.09 | .. | 0.0 |
| PIANO | play | 0.25 | 0.33 | 0.36 | 0.38 | .. | 0.5 |
| PIANO | piano | 0.25 | 0.33 | 0.36 | 0.38 | .. | 0.5 |
| PLAY | i | 0.25 | 0.16 | 0.12 | 0.09 | .. | 0.0 |
| PLAY | play | 0.25 | 0.33 | 0.36 | 0.38 | .. | 0.5 |
| PLAY | piano | 0.25 | 0.33 | 0.36 | 0.38 | .. | 0.5 |
| UNDERSTAND | i | 0.25 | 0.25 | 0.21 | 0.17 | .. | 0.0 |
| UNDERSTAND | understand | 0.25 | 0.50 | 0.55 | 0.61 | .. | 1.0 |

In IBM Model 2, we add an explicit model for alignment. In IBM Model 1, we do not have a probabilistic model for this aspect of translation. As consequence, according to IBM Model 1 the translation probabilities for the two examples cited previously are the same. IBM Model 2 addresses the issue of alignment with an explicit model for alignment based on the positions of the input and output words. The translation of an English input word in position i to an ASL word in position j is modeled by an alignment probability distribution:

$$a(i|j, l_e, l_f)$$

Recall that the length of the input sentence f is denoted as $l_f$, and the length of the output sentence e is $l_e$. We can view translation under IBM Model 2 as a two-step process with a lexical translation step and an alignment step:

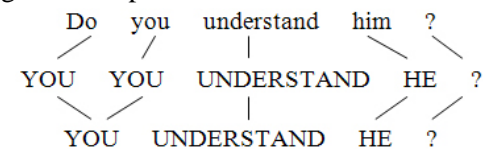

The first step is lexical translation as in IBM Model 1, again modeled by the translation probability t(e|f). The second step is the alignment step. For instance, translating 'understand' into 'UNDERSTAND' has a lexical translation probability of :

t(UNDERSTAND | understand)

and an alignment probability of a(2|3,4,5) - the 2$^{th}$ ASL word is aligned to the 3rd English word.

Note that the alignment function a maps each ASL output word j to an English input position a(j) and the alignment probability distribution is also set up in this reverse direction. The two steps are combined mathematically to form IBM Model 2:

$$p(e,a|f) = \epsilon \prod_{j=1}^{l_e} t(e_j|f_{a(j)}) a(a(j)|j, l_e, l_f)$$

Table 3 shows that in only three iterations we achieve the same results of IBM Model 1.

In IBM Model 3, we account the NULL token. In other words, you can get a word in English that is not translated into ASL. The probability of generating a NULL token is:

$$p(\emptyset_0) = \binom{l_e - \emptyset_0}{\emptyset_0} p_1^{\emptyset_0} p_0^{l_e - 2\emptyset_0}$$

Table 3. Application of IBM Model 2 EM Training

| e | f | initial | 1$^{st}$ it. | 2$^{nd}$ it. | 3$^{rd}$ it. |
|---|---|---|---|---|---|
| I | i | 0.64 | 0.73 | 0.96 | 1.00 |
| I | understand | 0.38 | 0.32 | 0.09 | 0.00 |
| I | play | 0.23 | 0.18 | 0.03 | 0.00 |
| I | piano | 0.23 | 0.18 | 0.03 | 0.00 |
| PIANO | i | 0.09 | 0.06 | 0.00 | 0.00 |
| PIANO | play | 0.38 | 0.40 | 0.48 | 0.50 |
| PIANO | piano | 0.38 | 0.40 | 0.48 | 0.50 |
| PLAY | i | 0.09 | 0.06 | 0.00 | 0.00 |
| PLAY | play | 0.38 | 0.40 | 0.48 | 0.50 |
| PLAY | piano | 0.38 | 0.40 | 0.48 | 0.50 |
| UNDERSTAND | i | 0.17 | 0.13 | 0.01 | 0.00 |
| UNDERSTAND | understand | 0.61 | 0.67 | 0.90 | 1.00 |

Due to the problem of incomplete and according to Koehn, we are facing a typical problem for machine learning. We want to estimate our model from incomplete data. So, we will use the expectation Maximization algorithm, or EM Algorithm that addresses the situation of incomplete data. It is an iterative learning method that fills in the gaps in the data and trains a model in alternating steps. We apply EM for IBM Model 1, 2 and 3.

## 6. String Matching

Words in American Sign Language are very similar to English written text. So, we think to use others techniques to learn data quickly and efficiency, for example, string-matching. String-matching is a very important subject in the wider domain of text processing. String-matching algorithms are basic components used in implementations of practical software existing under most operating systems. Moreover, they emphasize programming methods that serve as paradigms in other fields of computer science. They also play an important role in theoretical computer science by providing challenging problems. String-matching consists in finding one, or more generally, all the occurrences of a string in a text or with another string. The pattern is denoted by $x = x[0 .. m − 1]$; its length is equal to m. The text is denoted by $y = y[0 .. n − 1]$; its length is equal to n. Both strings are build over a finite set of character called an alphabet denoted by with size is equal to. Several algorithms and methods exist like Jaro-Winkler distance that does will be used in word alignment process from statistical sign language machine translation.

6.1 Jaro-Winkler distance

The Jaro–Winkler distance [17] is a measure of similarity between two strings. It is a variant of the Jaro distance metric and mainly used in the area of record linkage. The higher the Jaro–Winkler distance for two strings is, the more similar the strings are. The Jaro–Winkler distance metric is designed and best suited for short strings such as person names. The score is normalized such that 0 equates to no similarity and 1 is an exact match. The Jaro distance $d_j$ of two given strings $S_1$ and $S_2$ is:

$$d_j = \frac{1}{3}\left(\frac{m}{|s_1|} + \frac{m}{|s_2|} + \frac{m-t}{m}\right)$$

Where:
- m is the number of matching characters
- t is the number of transpositions

Jaro–Winkler distance uses a prefix scale $p$ which gives more favorable ratings to strings that match from the beginning for a set prefix length l. Given two strings $S_1$ and $S_2$, their Jaro–Winkler distance $d_w$ is: $d_w = d_j + \left(l.p.(1 - d_j)\right)$

Where:
- $d_j$ is the Jaro distance between $s_1$ and $s_2$.
- l is the length of common prefix at the start of the string up to maximum of 4 characters.
- p is constant scaling factor for how much the score is adjusted upwards for having common prefixes. p should not exceed 0.25, otherwise the distance can become larger than 1. The standard value for this constant in Winkler's work is p = 0.1.

The next Table presents some examples:

Table 4. Jaro-Winkler distance applied to 5 pairs word

| S1 | S2 | Jaro distance | Jaro-Winkeler distance |
|---|---|---|---|
| I | i | 1.00 | 1.0000 |
| I | understand | 0.00 | 0.0000 |
| PIANO | play | 0.38 | 0.4550 |
| PIANO | piano | 1.00 | 1.0000 |
| UNDERSTAND | understand | 1.00 | 1.0000 |

6.2 String Matching for EM for IBM Model 1

Starting from the formula: $p(a|e, f) = \frac{p(e, a|f)}{p(e|f)}$

We improve the result by adding the $d_w$ between e and f, we have:

$$p(a|e, f) = \frac{\alpha.p(e, a|f) + (1-\alpha).d_w(e, f)}{p(e|f)}$$

Where α is the coefficient of similiraty between the two words e and f. The standard value of α used for experiments is 0.5. Table 2 presents comparative results applied to a small corpus composed by two pair-sentences.

Table 5. Application of IBM Model 1 EM Training with string matching

| e | f | initial | 3 iterations | 3 iterations + String Matching |
|---|---|---|---|---|
| I | i | 0.2500 | 0.6412 | 0.9684 |
| I | understand | 0.2500 | 0.3879 | 0.0532 |
| I | play | 0.2500 | 0.2307 | 0.0316 |
| I | piano | 0.2500 | 0.2307 | 0.0839 |
| PIANO | i | 0.2500 | 0.0929 | 0.0475 |
| PIANO | play | 0.2500 | 0.3846 | 0.3738 |
| PIANO | piano | 0.2500 | 0.3846 | 0.7977 |
| PLAY | i | 0.2500 | 0.0929 | 0.0049 |
| PLAY | play | 0.2500 | 0.3846 | 0.8170 |
| PLAY | piano | 0.2500 | 0.3846 | 0.3741 |
| UNDERSTAND | i | 0.2500 | 0.1727 | 0.0123 |
| UNDERSTAND | understand | 0.2500 | 0.6120 | 0.9467 |

6.3 String Matching for EM for IBM Model 2

Table 6. Application of IBM Model 2 EM Training with string matching

| e | f | initial | 2 iterations | 2 iterations + string matching |
|---|---|---|---|---|
| I | i | 0.6412 | 0.7343 | 0.9986 |
| I | understand | 0.3879 | 0.3258 | 0.0030 |
| I | play | 0.2307 | 0.1899 | 0.0010 |
| I | piano | 0.2307 | 0.1899 | 0.0380 |
| PIANO | i | 0.0929 | 0.0657 | 0.0345 |
| PIANO | play | 0.3846 | 0.4050 | 0.3169 |
| PIANO | piano | 0.3846 | 0.4050 | 0.9047 |
| PLAY | i | 0.0929 | 0.0657 | 0.0000 |
| PLAY | play | 0.3846 | 0.4050 | 0.9044 |
| PLAY | piano | 0.3846 | 0.4050 | 0.3129 |
| UNDERSTAND | i | 0.1727 | 0.1341 | 0.0001 |
| UNDERSTAND | understand | 0.6120 | 0.6741 | 0.9969 |

Like to IBM Model 1, we add an α coefficient for string matching to alignment process. Results show that we converge to 1 after 2 iterations only in a small corpus. We note that the corpus contains two similar words but have not the same semantic and role 'piano' and 'play'. The next table presents the experimental results.

## 7. Phrase-based model and Decoding

In phrase-based translation, the aim is to reduce the restrictions of word-based translation by translating whole sequences of words, where the lengths may differ. We use MOSES tool to learn phrase alignment. After that, we exploit the decoding tool. This step is the main function in the system. The input is an English sentence. The role of the decoder is to find the best translation. The probabilistic model for phrase-based translation is:

$$e_{max} = argmax_e \prod_{i=1}^{l} \emptyset(\overline{f_i}|\overline{e_i})d(start_i - end_{i-1} - 1)P_{LM}(e)$$

Where:
- Phrase translation: picking phrase $\overline{f_i}$ to be translated as a phrase $\overline{e_i}$. We look up score $\emptyset(\overline{f_i}|\overline{e_i})$ from phrase translation table.
- Reordering: Previous phrase ended in $end_{i-1}$, current phrase starts at $start_i$. We compute $d(start_i - end_{i-1} - 1)$.
- Language Model : For n-gram model, we need to keep track of last $(n-1)$ words. We compute score $P_{LM}(W_i|W_{i-(n-1)},\ldots,W_{i-1})$ for added words $W_i$.

## 8. Results and Word alignment matrix

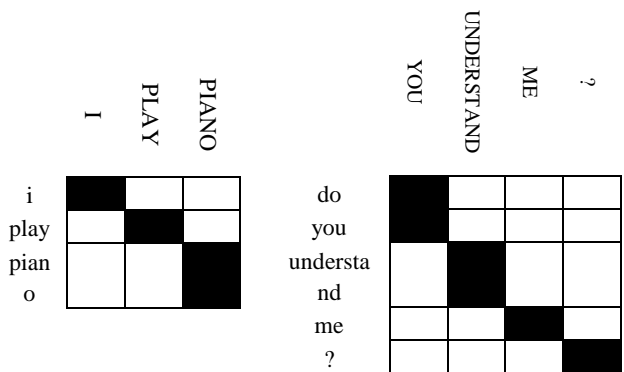

Figure 3. Word alignment matrix: Words in the ASL (columns) are aligned to words in the English sentence (rows) as indicated by the filled points in the matrix

One way to visualize the task of word alignment is by a matrix as in Figure 3. Here, alignments between words (for instance between the English 'play' and the ASL 'PIANO') are represented by points in the alignment matrix. Word alignments do not have to be one-to-one. Words may have multiple or no alignment points. For instance, the ASL word assumes is aligned to the two English words 'do you'. However, it is not always easy to establish what the correct word alignment should be. Experimentation and results

In this evaluation, we trained a small 3-gram language model using data in Table 7.

Results are very encouraged. Table 8 shows some alignment sentences with scores.

For interpretation, we use WebSign tool [18]. WebSign is a project that carries on developing tools able to make information over the web accessible for deaf.

Table 7. Statistics of Parallel data

| Language | Sentences | Tokens |
|---|---|---|
| English | 431 | 632 |
| ASL | 431 | 608 |
| n-gram 1 = 609 - n-gram 2 = 1539 - n-gram 3 = 257 | | |

## 9. Conclusions

We describe several experiments with English-to-American Sign Language statistical sign language machine translation. Employing a technique of string matching is crucial. In conclusion, phrase-based statistical MT for sign language from English to American Sign Language performs well, despite the expectations arising from linguistic knowledge about the properties of ASL. This work we experimented with is currently the best performing machine translation evaluated on this pair of languages.

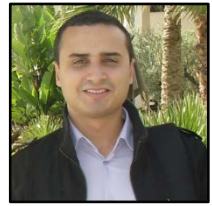

**Achraf Othman** is currently a PhD student under the supervision of Prof. Mohamed Jemni. He received in August 2010 the Master degree on Computer Science from Tunis College of Sciences and Techniques (ESSTT), University of Tunis in Tunisia. His research interests are in the areas of Sign Language Processing. His current topics of interests include Grid Computing, Computer graphics and Accessibility of ICT to Persons with Disabilities.

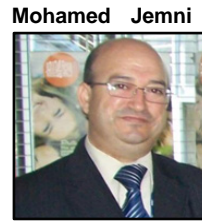

**Mohamed Jemni** is a Professor of ICT and Educational Technologies at the University of Tunis, Tunisia. He is the Head of the Laboratory Research of Technologies of Information and Communication (UTIC). Since August 2008, he is the General chair of the Computing Center El Khawarizmi, the internet services provider for the sector of the higher education and scientific research. His Research Projects Involvement are tools and environments of e-learning, Accessibility of ICT to Persons with Disabilities and Parallel & Grid Computing.

Table 8. Some alignments sentences (English / American Sign Language) with scores

| ASL Sentence pair<br>English sentence + alignment | Length :<br>Source / Target | Score |
| --- | --- | --- |
| DEAF YOU ?<br>NULL ({ }) are ({ }) you ({ 2 }) deaf ({ 1 }) ? ({ 3 }) | Source : 4<br>Target : 3 | 0.0016781 |
| YOU UNDERSTAND SHE , YOU ?<br>NULL ({ 4 }) do ({ 5 }) you ({ 1 }) understand ({ 2 }) her ({ 3 }) ? ({ 6 }) | Source : 5<br>Target : 6 | 5.387e-07 |
| YOU FAVORITE- [ prefer ] , HAMBURGER [ body-shift-or ] HOTDOG ?<br>NULL ({ 6 }) do ({ }) you ({ 1 }) prefer ({ 2 5 10 }) hamburgers ({ 7 9 }) or ({ 3 8 }) hotdogs ({ 4 11 }) ? ({ 12 }) | Source : 7<br>Target : 12 | 2.195e-16 |
| last-YEAR TICKET HOW-MANY YOU ?<br>NULL ({ }) how ({ }) many ({ 3 }) tickets ({ 1 2 }) did ({ }) you ({ 4 }) get ({ }) last ({ }) year ({ }) ? ({ 5 }) | Source : 9<br>Target : 5 | 3.661e-06 |
| TOPIC YOU DON 'T-CARE WHAT ?<br>NULL ({ }) what ({ }) do ({ }) you ({ 2 }) not ({ 3 }) care ({ 1 4 }) about ({ 5 }) ? ({ 6 }) | Source : 7<br>Target : 6 | 5.444e-06 |
| DRESS YOU LIKE USE- [ wear ] YOU ?<br>NULL ({ }) do ({ 8 }) you ({ 2 }) like ({ 3 }) to ({ }) wear ({ 1 4 6 }) dresses ({ 5 7 }) ? ({ 9 }) | Source : 8<br>Target : 7 | 2.799e-11 |
| WET-WIPES YOU KEEP CAR ?<br>NULL ({ }) do ({ }) you ({ 2 }) keep ({ }) wet ({ 1 }) wipes ({ 3 }) in ({ }) your ({ }) car ({ 4 }) ? ({ 5 }) | Source : 9<br>Target : 5 | 4.286e-05 |
| STRIPES- [ vertical ] , YOU FACE- [ look ] GOOD YOU ?<br>NULL ({ }) do ({ 6 }) you ({ 12 }) look ({ 4 7 10 }) good ({ 2 8 11 }) in ({ 5 }) stripes ({ 1 3 9 }) ? ({ 13 }) | Source : 7<br>Target : 13 | 6.651e-19 |
| # Sentence pair (430) source length 6 target length 4 alignment score :<br>WHAT YOU ENTHUSIASTIC ?<br>NULL ({ }) what ({ }) are ({ }) you ({ 2 }) enthusiastic ({ 3 }) about ({ 1 }) ? ({ 4 }) | Source : 6<br>Target : 4 | 0.0006027 |